\documentclass[a4paper,conference]{IEEEtran}
\usepackage{graphicx}

\ifCLASSINFOpdf
\else
\fi
\usepackage{amsmath}
\usepackage{amssymb}
\usepackage{array}
\usepackage{bm}
\begin{document}
%
\title{Precision Learning: Towards Use of Known Operators in Neural Networks}

\author{\IEEEauthorblockN{Andreas Maier$^*$, Frank Schebesch$^*$, Christopher Syben$^*$, Tobias W{\"u}rfl$^*$, Stefan Steidl$^*$, \\Jang-Hwan Choi$^+$, Rebecca Fahrig$^*$}
\IEEEauthorblockA{$*$ Pattern Recognition Lab\\
Friedrich-Alexander-University Erlangen-Nuremberg, Germany\\
$+$ Medical Imaging Systems Lab, Division of Mechanical and Biomedical Engineering,\\
College of Engineering, Ewha Womans University, Seoul 03760, Korea\\
Email: andreas.maier@fau.de}
}


%


\maketitle

\begin{abstract}
In this paper, we consider the use of prior knowledge within neural networks. In particular, we investigate the effect of a known transform within the mapping from input data space to the output domain. We demonstrate that use of known transforms is able to change maximal error bounds.

In order to explore the effect further, we consider the problem of X-ray material decomposition as an example to incorporate additional prior knowledge. We demonstrate that inclusion of a non-linear function known from the physical properties of the system is able to reduce prediction errors therewith improving prediction quality from SSIM values of 0.54 to 0.88.

This approach is applicable to a wide set of applications in physics and signal processing that provide prior knowledge on such transforms. Also maximal error estimation and network understanding could be facilitated within the context of \textit{precision learning}.
\end{abstract}


%
\IEEEpeerreviewmaketitle

\section{Introduction}
Neural networks and deep learning recently produced numerous exciting results in pattern recognition and machine intelligence \cite{AlphaGo,YOLO,ghesuDL,aubreville}. Most of these approaches use hand-crafted deep network architectures and all weights are trained starting from a random initialisation. One effect of this procedure is that for some applications hundreds of millions of training samples are required to get optimal performance \cite{ImageNet,googlepaper}. Interestingly, emerging networks share similarities with known operators \cite{sindhwani15, humphrey2012moving}. These approaches seem adequate for general perceptive tasks as prior knowledge is difficult to obtain. Doing so gives rise to the current trend of ``network engineering''. However, in many slightly more restricted application problems, a broad body of prior knowledge is available, e.g. physics, signal processing, and other expert knowledge.

\begin{figure*}[h!]
\begin{center}
\includegraphics[width=0.9\linewidth]{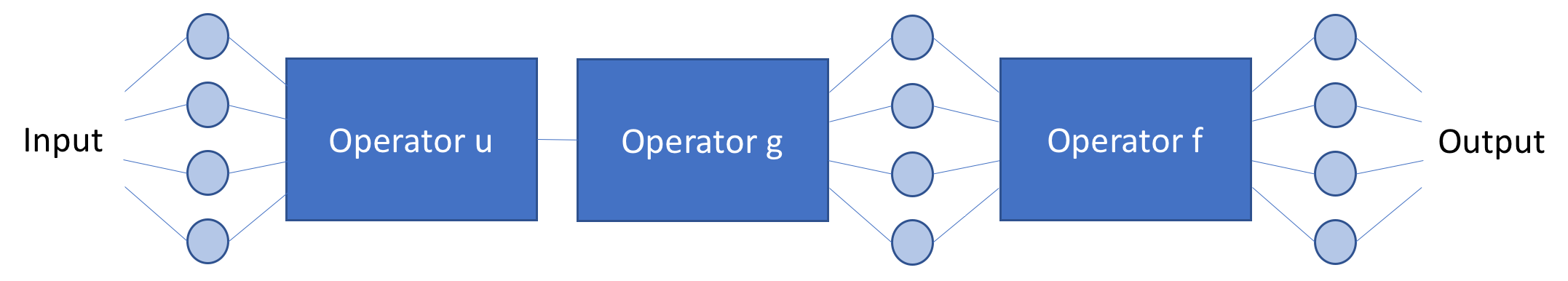}
\caption{Schematic of the idea of precision learning: One or more known operators are embedded into a network. Doing so, allows dramatic reduction of the number of parameters that have to be estimated during the learning process. The minimal requirement for the operator is that it must allow the computation of a sub-gradient for use in combination with the back-propagation algorithm. This requirement can be met by a large class of operations.}
\label{fig:precision}
\end{center}
\end{figure*}

Only recently a few investigations using known operators in the neural network modelling process have been published. W{\"u}rfl et al. demonstrated that back-projection and filtering of CT reconstruction can be embedded into a neural network context allowing the learning of redundancy weights \cite{deeplearningct}. Based on these observations Syben et al. showed that neural networks are also able to learn discrete filter versions from their continuous counterparts \cite{ISBIArchiveSyben}. By design the network is restricted to a general reconstruction filter and by clever synthesis of training data, 10 observations are sufficient for the learning process. Fu et al. developed this concept even further and demonstrated that even more complex operations such as the eigenvalue computation in Frangi's vesselness can be modelled with mathematical equivalence in a neural Network design \cite{ArXivWeilin}. As all of these approaches fix parts of the network due to prior information about the underlying algorithm, the number of unknown parameters is dramatically reduced. Still neural Network training is employed using typical optimisation approaches as provided by libraries such as TensorFlow. Additionally they employ specialised layers that are able to compute the gradient of the network efficiently. We consider such approaches that mix known functions with unknown operators as examples of {\it precision learning} which typically requires less training examples and fewer training iterations. Figure~\ref{fig:precision} shows a graph of such an approach conceptually.

In this paper, we wish to explore the general bounds on such approaches and make comparisons to the universal approximation theorem \cite{cybenko1989approximation}. Based on these observations, we pick an example in which machine learning has been successfully applied previously: X-ray material decomposition \cite{univis91549326, zimmerman2015experimental}. We briefly investigate its theory and explore the effect of using known operators on the quality of the material decomposition to verify our theoretical observations.

\section{Bounds for Sequences of Operators}
\newcommand{\x}{{\bf x}}
\newcommand{\uu}{{\bf u}}
\newcommand{\UU}{{\bf U}}
\renewcommand{\d}[1]{\ensuremath{\operatorname{d}\!{#1}}}

\begin{figure*}[tb]
\begin{center}
\includegraphics[width=0.4\linewidth]{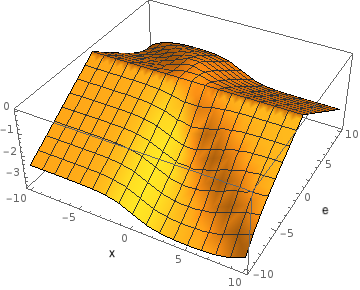}\hspace{1cm}\includegraphics[width=0.4\linewidth]{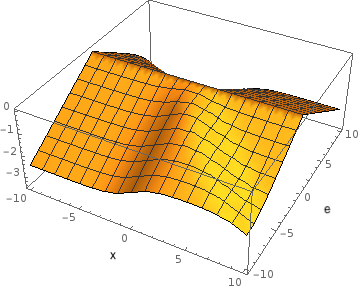}
\caption{Plots for the relation $g_j s(x+e) - g_j s(x) - |g_j|\cdot l_s \cdot |e| \le 0$ (cf. Eq.~\ref{eq:upperbound}) for $g_j>0$ on the left hand side and for $g_j <0$ on the right hand side, normalised by $|g_j|$. As can be seen in the plots, the inequality holds for both cases.}
\label{fig:bound}
\end{center}
\end{figure*}

The universal approximation theorem proves that any continuous function $u(\x)$ on a compact subset of $\mathbb{R}^N$ can be approximated by a linear combination of sigmoid functions $s(x)$
\begin{equation}
u(\x) \approx U(\x) = \sum_i u_i s({\bf w}^\top {\bf {x}} +w_0)
\end{equation}
where ${\bf w}$ and $w_0$ are hyper plane parameters that project $\x$ and $u_i$ are the weights of the linear superposition of the sigmoid curves. The resulting approximation error $\epsilon_u$ is an upper bound following 
\begin{equation}
|U(\x) - u(\x)| \le \epsilon_u.
\end{equation}
Following this concept, we can expand to a multidimensional $\UU(\x)$ and $\uu(\x)$ in which every component follows 
\begin{equation}
|U_j(\x) - u_j(\x)| \le \epsilon_{u_j}.
\end{equation}
This enables us to explore the chain of operators $f(\x)=g(\uu(\x))$. In the following, we will refer to the approximations of the original functions as capital letters,\,i.e., $G(\x)$ is the approximation of $g(\x)$. Furthermore, we link the plane projection ${\bf w}^\top {\bf {x}} +w_0$ only to the input layer which gives rise to a typical hidden layer network structure with alternating linear combinations and sigmoid activation functions. This leads to the following possible approximations
\begin{eqnarray}
F_u(\x)=&g(\UU(\x)) &= f(\x)-e_u\\
F_g(\x)=&G(\uu(\x))&= f(\x)-e_g\\
F(\x)=&G(\UU(\x)) &= f(\x)-e_f \label{eq:ef}
\end{eqnarray}
where $F_u(\x)$ uses the approximation of $\uu(\x)$, but knows $g(\x)$, $F_g(\x)$ knows $\uu(\x)$, but approximates $g(\x)$, and $F(\x)$ needs to approximate both. Respectively, $e_u$, $e_g$, and $e_f$ are the errors that are introduced by these approximations. Each of these errors is bound by a positive corresponding $\epsilon$,\,i.e., $|e_u| \le \epsilon_u$. Of particular interest is the chain of operators
\begin{equation}
F(\x) = G(\UU(\x))= \sum_j g_j s\left(\sum_i u_{i,j} s({\bf w}^\top {\bf {x}} +w_0)\right) + g_0
\end{equation}
and bounds to this approximation. Based on our previous observations, we find
\begin{eqnarray}
f(\x)&=& g(\uu(\x))= G(\uu(\x)) +e_g \nonumber\\
&=&\sum_j g_j s\left(u_j(\x)\right)  + g_0 + e_g\nonumber\\
&=& \sum_j g_j s\left(U_j(\x) + e_{u_j}\right)  + g_0 + e_g. \label{eq:defF}
\end{eqnarray}
If we consider the case of overestimation,\,i.e. $f(\x) \le F(\x)$,
we are interested in finding an upper bound for the approximation.
As a broad range of functions, the sigmoid function is Lipschitz continuous,\,i.e. its slope is bounded by a positive value $l$. For the sigmoid function, this value is found at $s(0)$ with $l_s=0.25$. Thus, a general upper bound for $s(x+e)$ is
$$s(x+e) \le s(x) + l_s \cdot |e|.$$
When combined with a multiplicative constant $g_j$, this upper bound needs to be modified
to
\begin{equation}
g_j s(x+e) \le g_j s(x) + |g_j| \cdot l_s \cdot |e|. \label{eq:upperbound}
\end{equation}
This relation is visualised in Figure~\ref{fig:bound}.
Hence, we can find an upper bound of Eq.~\ref{eq:defF} as
\begin{eqnarray}
f(\x)\le& \underbrace{\sum_j g_j s(U_j(\x)) + g_0}{}+ \sum_j |g_j| \cdot l_s \cdot |e_{u_j}|  + e_g\nonumber\\
\le& F(\x) + \sum_j |g_j| \cdot l_s \cdot |e_{u_j}| + e_g.
\end{eqnarray}
Subtraction of $F(\x)$ yields
\begin{eqnarray}
\underbrace{f(\x) - F(\x)} &\le&  \sum_j |g_j| \cdot l_s \cdot |e_{u_j}| + e_g\nonumber\\
e_f ~~~~~~&\le&  \sum_j |g_j| \cdot l_s \cdot |e_{u_j}| + e_g.
\end{eqnarray}
Using $|e_g| \le \epsilon_g$ leads to
\begin{equation}
e_f\le \sum_j |g_j| \cdot l_s \cdot |e_{u_j}| + \epsilon_g.
\end{equation}

For the case of underestimation with $f(\x) \ge F(\x)$, a similar
bound is found as
\begin{equation}
e_f \ge - \sum_j |g_j| \cdot l_s \cdot |e_{u_j}| - \epsilon_g
\end{equation}
Thus, an absolute bound for $e_f$ is found as
\begin{equation}
|e_f| \le \sum_j |g_j| \cdot l_s \cdot |e_{u_j}| + \epsilon_g.
\label{eq:ef}
\end{equation}
This observation is interesting as $\sum_j |g_j| \cdot l_s \cdot |e_{u_j}|$ is also an upper bounded
for $|e_u|$, if $g(\x)$ is a superposition of sigmoid functions. If we consider a more general $g(\x)$,
bound by the Lipschitz constant $l_g$, we obtain the following upper bound
\begin{eqnarray}
g(\x +{\bf e_u} ) \le g(\x) + l_g \cdot {|| {\bf e_u}||}_1
\end{eqnarray}
where ${\bf e_u}$ is a vector containing all $e_{u_j}$ and ${||\cdot||}_1$ is the L$_1$ norm. Following the derivation above, it is easy to see, that 
\begin{eqnarray}
|e_u| \le& l_g \cdot {|| {\bf e_u} ||}_1.
\end{eqnarray}
Conceptually both bounds are similar as
\begin{eqnarray}
 l_g \cdot {||{\bf e_u}||}_1 = \sum_j l_g \cdot |e_{u_j}| \approx \sum_j |g_j| \cdot l_s \cdot |e_{u_j}|.
\end{eqnarray}
This means that $|e_f|$ is essentially bounded by $\epsilon_u$ and $\epsilon_g$.
Thus, the maximal estimation error achieved using known operators
is lower or equal to the error achieved when training $F(\x)$ from scratch without prior knowledge. Knowledge both on $g(\x)$ and $\uu(\x)$ respectively reduces the maximal error in an additive fashion which leads us to the conclusion that any use of prior domain knowledge is useful to support the learning problem.
This forms a theoretical foundation for the success of previous applications of {\it precision learning}.

\begin{figure*}[tb]
\begin{center}
\includegraphics[width=\linewidth]{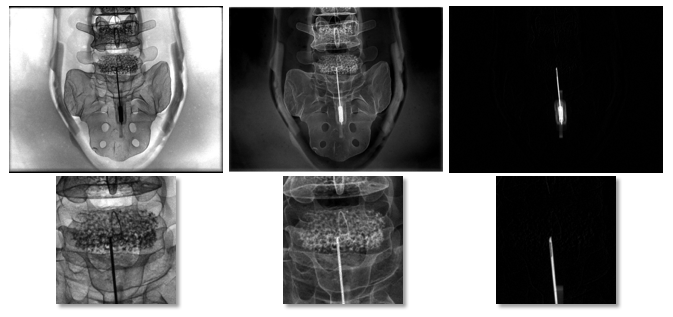}
\caption{Example images of the different data: In the left column, an X-ray image $I_b(x,y)$ of the phantom and the biopsy needle are shown. The center images show the data after transform $u(I_b(x,y))$,\,i.e. line integral domain. The right column shows the subtracted image after needle removal that was used as ground truth. Note that the needle grip is embedded in plastic which is of course a different material than metal.}
\label{fig:x-ray}
\end{center}
\end{figure*}

\section{X-Ray Material Decomposition}
In this section, we want to explore an example for Eq.~\ref{eq:ef}. In order to do so, we choose X-ray material decomposition as many properties of the transform are known. Yet analytical inversion requires accurate calibration of the X-ray spectrum $I_0(E)$ where $E$ is the X-ray energy. For each energy,
the line integral along the X-ray beam is observed
\begin{equation}
\displaystyle
I(E)=I_0(E) \cdot \mathrm{e}^{-\sum_i \mu(i,E) \l_i}
\end{equation}
where $\mu(i,E)$ is the X-ray absorption coefficient for energy $E$ and material $i$ and $l_i$ is the path length through material $i$. Typically, X-ray detectors measure the integral over several energies
\begin{equation}
I_b(x,y) = \int_{b_l}^{b_u} I(x, y, E)\, \mathrm{d}E
\end{equation}
at each pixel $(x,y)$ from the low energy limit $b_l$ to the upper energy limit $b_u$ of the respective energy bin $b$. Inversion can be performed iteratively, if the spectrum and the material coefficients are accurately known \cite{univis91424501}. In practice, this process is difficult as generally all materials in the object under consideration have to be known and the spectral calibration is error prone. A common approach is to model the decomposition using a low-order polynomial \cite{maass2011empirical} as the function is generally smooth. Lu et al. demonstrated that this approach is a general regression problem that predicts $l_i$ given an approximation of the inverse function $F({\bf I})$ where $\bf I$ is the vector containing all measured energy bins \cite{univis91549326}. Amongst several machine learning techniques, multi-layer perceptrons demonstrated the best results. Lu et al. already used a logarithmic transform on their data and expanded the coefficients to a polynomial space as they knew that both assumptions are supposed to be part of the regression problem. 

\section{Experimental Results}

In order to explore our idea, we will introduce a $-\log$ transform given the calibrated $I_{0}$ in each energy bin for $\uu(\x)$ and a polynomial expansion for $g(\x)$ following the ideas presented in \cite{maass2011empirical}. The remaining estimation problem is left to a three-layer perceptron $F({\x})$ which has a higher modelling ability than a single hidden layer network. We explore the effect of the transforms on the following four combinations: $F({\bf I})$, $F(u({\bf I}))$, $F(g({\bf I}))$, and $F(g(u({\bf I})))$. All implementations were performed in the CONRAD open source software package \cite{conradpaper}.

The experimental data for our experiment was collected at Stanford University using a Zeego C-arm device (Siemens Healthineers, Forchheim). Using prototype software, scans of a Sawbones pelvis phantom were acquired. For the three different energy settings, 41, 70, and 125kVP were selected. Image dimensions were $620\times480$ pixels. In order to get reference labels for the ground truth, we inserted a biopsy needle into the phantom, performed the three scans, and carefully removed the needle to cause minimal motion of the phantom. Subsequently, another series of three scans was performed. In order to generate the ground truth, the 125~kVp scans before and after needle insertion were subtracted. Examples of the X-ray images are shown in Figure~\ref{fig:x-ray}.

\begin{table}[htp]
\caption{Overview on the results of the prediction. Pearson's $r$ is generally high, while the SSIM is drastically increased with increasing prior knowledge.}
\begin{center}
\begin{tabular}{|l|c|c|c|c|}
\hline
 & $F({\bf I})$&$F(u({\bf I}))$&$F(g({\bf I}))$&$F(g(u({\bf I})))$\\
 \hline
 Pearson's $r$ [\%]& 95.0 & 95.2 & 95.1& {\bf95.5}\\
 SSIM [\%] & 54.1 & 63.1 & 73.8 & {\bf88.4}\\
 \hline
\end{tabular}
\end{center}
\label{tab:res}
\end{table}%

In total, training and test contained 273.360 feature vectors. We evaluated the prediction quality using Pearson's $r$ \cite{pearson1895correlation} and the Structural Similarity Index (SSIM) \cite{wang2004image}. Table~\ref{tab:res} gives an overview on the results. For both correlation as well as SSIM, the prediction quality increases with increasing use of prior knowledge.

\begin{figure}[tb]
\begin{center}
\includegraphics[width=\linewidth]{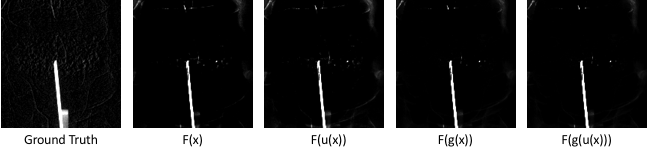}
\caption{Even in narrow window and level [0, 0.3], the material separation generally works well. Note that in particular, the artefacts in the background gradually disappear with increasing use of prior knowledge. In addition, note that the plastic grip in the bottom of the image generally disappears in the
predicted data, as the grip does not consist of metal.}
\label{fig:pred}
\end{center}
\end{figure}

Figure~\ref{fig:pred} shows the prediction result visually. The needle is visualised well in all four versions of the material decomposition and the plastic grip is removed. No bone is shown in the image anymore. Still, there are some artefacts in the background of the image that decrease with increasing use of prior knowledge.

\begin{figure*}[tb]
\begin{center}
\includegraphics[width=0.5\linewidth]{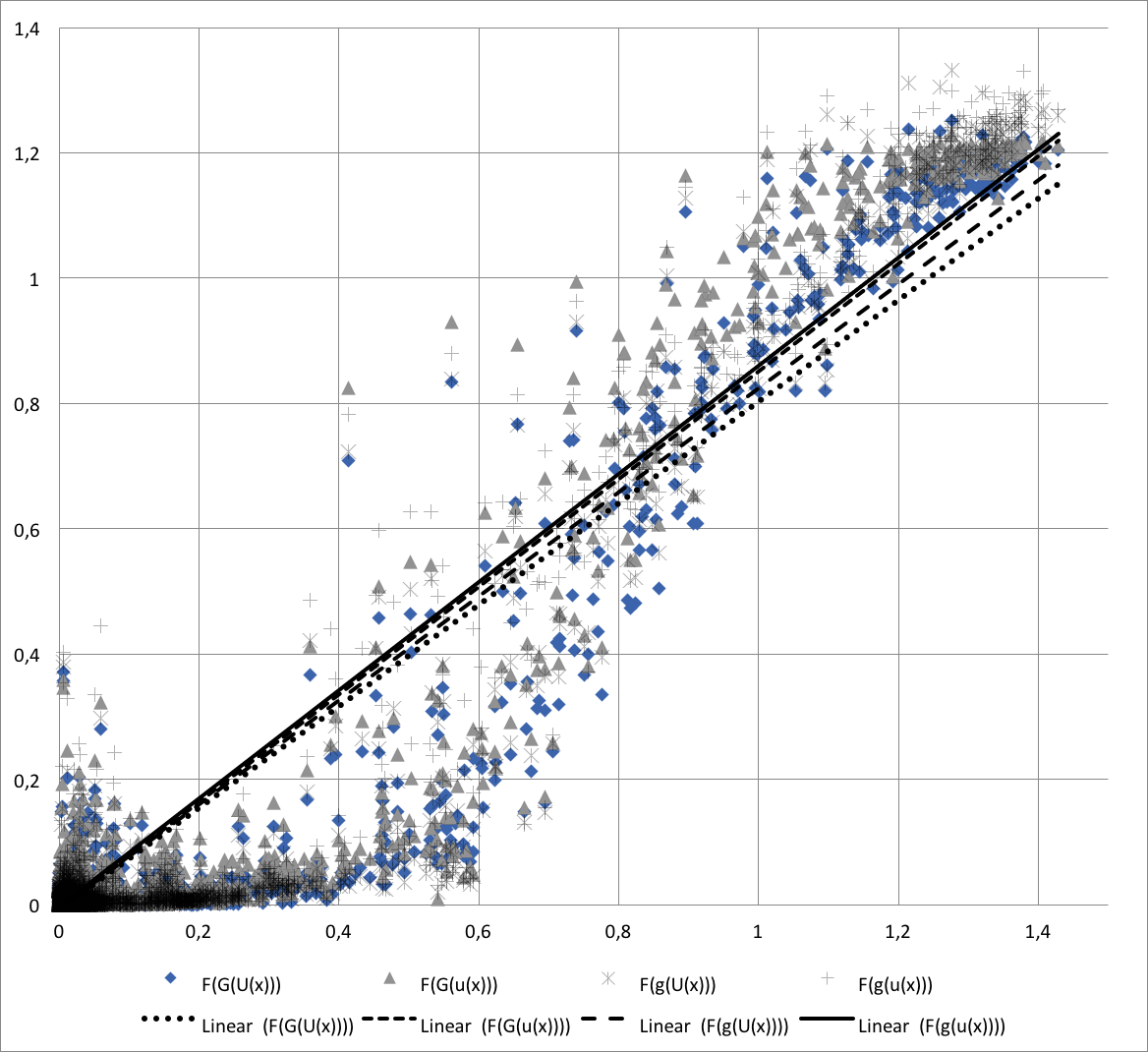}
\caption{Display of the scatter plot ground truth vs. prediction of the four predictors. Note that the shape of the scattering becomes more and more linear with increasing use of prior knowledge.}
\label{fig:plot}
\end{center}
\end{figure*}

In Figure~\ref{fig:plot}, we show the estimation vs. the ground truth of all four methods in comparison. With increasing use of prior knowledge,\,i.e., transforms $\uu(\x)$ and $g(\x)$, the prediction becomes more and more linear. Also note that the plastic in the reference introduces a bias in the estimation of the low path lengths through the needle. 

\section{Discussion}

In this paper, we show an upper bound for errors that are introduced by the approximation of functions using neural networks. In our derivation, we follow the universal approximation theorem using a single layer per function. Generally, the approximation introduces an error and a combination of multiple approximations increases the upper error bound. This approach could also be expanded to more layers. Using our approximation formula, this would only increase the maximal bound further. Hence, we consider one layer as demonstrated in the universal approximation theorem as sufficient. Furthermore, it is noteworthy to state that the universal approximation theorem always combines one hyper plane with one weighted superposition of sigmoid functions. Here, we only do that in the input layer, as we wanted our model to mimic traditional hidden layer approaches. Additional use of another hyper plane in the second function approximation would not change our results as this is already modelled in the linear combination of ${\bf U}(\x)$. Yet, the notation would become less tangible. Therefore, we chose the current notation.

Our upper bounds share further similarities with the universal approximation theorem: Specification of the number of nodes per layer is not required and with increasing number of nodes, the maximal bounds for the error will shrink. Furthermore, we also do not show that our upper bound is tight. It is merely a worst-case analysis and does not describe the average or expected error. Further analysis could be performed based on the work of Barron \cite{barron1994approximation}.

Still, these observations are useful for understanding other applications of {\it precision learning} that already have been published \cite{deeplearningct,ISBIArchiveSyben,ArXivWeilin}. With the bounds found in this paper, we are able to see that the maximal error margin decreases, independent of whether the transform is applied before or after the network, if that is the true sequence of operators. This confirms experimental findings that replace operators within the network to increase the prediction quality \cite{ArXivWeilin}.

Using this framework enables us to mix operators with deep learning procedures and provides a new view and understanding of deep learning. The benefits are at least two-fold: First, the mixing allows us to reduce the number of parameters of the network and therewith a reduction in training data \cite{ISBIArchiveSyben}. Second, it opens the view on neural networks as an ensemble of operators that we work with in terms of mathematical analysis. Technically, this also enables to ``derive'' new networks. Even operators that are typically more difficult to grasp in a mathematical sense can be used in this framework if they allow the computation of sub-gradients since the gradients can be processed in the back-propagation chain. This is also in line with our assumptions on $g(\x)$ being Lipschitz continuous. This even allows inclusion of sorting operators such as quantile operations \cite{miccai:schirrmacher}. As such this could potentially be a game changer for deep learning.

Furthermore,  {\it precision learning} will also help to understand complicated network structures in more detail. As such, we can now try to interpret the U-Net \cite{ronneberger2015u} algorithm as a multi-scale dictionary learning approach with learned shrinkage operators in which the reconstruction dictionary is independent of the analyser dictionary. Similar observations were already done for CNNs in \cite{papyan2016convolutional}. Hence, one could attempt to approximate its structure using traditional iterative shrinkage methods, wavelet and custom bases, and therewith reduce the number of parameters drastically.

The distinction of {\it precision learning} to general regularisation and deep learning is that we use precise domain knowledge expressed in terms of mathematical expressions to model our network instead of heuristically discovered network architectures. As such we interpret networks as a method to express a general objective function that describes an optimisation problem. Doing so, every block in the emerging objective function serves a particular purpose to which it is restricted to by design. Hence, the trained block can also be interpreted with respect to its use.

There are also links to classical pattern recognition theory \cite{niemann2013pattern}, if we interpret $\uu(\x)$ as the feature extractor and $g(\x)$ as the classifier/regressor. In particular, the observation that the error caused by $F_u(\x)$ has much higher bounds than the error produced by $F_g(\x)$ might also be linked to the classical importance of feature extraction and feature engineering. Even the best classifier cannot recover information lost during feature extraction.

In our experiments, we demonstrate another application of {\it precision learning}: X-ray material decomposition. We demonstrate that the use of additional prior domain knowledge is able to reduce the prediction error. As such, we only explore feature domain transformation in this paper. Given the realistic setting of our experiment, we deemed this to be more convincing. In every case, the material decomposition that is learned is physically correct and the flawed ground truth is mapped to a physically plausible result. With a constrained simulation experiment, we would not have been able to observe this finding.

\section{Conclusion \& Outlook}
We showed that {\it precision learning} is able to reduce maximal error bounds, if known operators are introduced into the learning process. This forms a theoretical basis for prior observations using the same approach. Our derivation shows a relationship between the number of approximation steps in their upper bounds. If replaced with the true transform this error is removed from the learning process. 

In addition, we explored the effect also in real data for an X-ray material decomposition and confirmed prior work by Lu et al. The more prior knowledge that is introduced, the better the prediction gets. In our case improvements from 54.1\,\% to 88.4\,\% in SSIM could be obtained.

We believe that this approach will be applicable to a large set of problems in particular in physics and signal processing where a lot of prior constraints on the domain are known. Also medical or other high risk applications could also benefit from this approach, as known operators in conjunction with few network layers are easier to analyse in terms of maximal error. Lastly, this approach also allows to design networks that follow a particular algorithm that enforces desired criteria, e.g. safety bounds on a separate path through the network.






\bibliographystyle{IEEEtran}
\bibliography{IEEEabrv,References}
%

\end{document}